\pdfoutput=1

\documentclass[11pt]{article}

\usepackage{ACL2023}

\usepackage{times}
\usepackage{latexsym}

\usepackage[T1]{fontenc}

\usepackage[utf8]{inputenc}

\usepackage{microtype}

\usepackage{inconsolata}

\usepackage{tikz}
\usepackage{times}
\usepackage{soul}
\usepackage{url}
\usepackage[utf8]{inputenc}
\usepackage{graphicx}
\usepackage{amsmath,amssymb}
\usepackage{amsthm}
\usepackage{booktabs}
\usepackage{algorithm}
\usepackage{algorithmic}
\usepackage{float}
\usepackage{fdsymbol}

\graphicspath{{figures/}}
\usepackage{subfig}
\usepackage{multirow}
\usepackage{bm}
\usepackage{color}
\definecolor{Gray}{gray}{0.935}

\newcommand{\xhdr}[1]{\vspace{1.6mm}\noindent{{\bf #1.}}}
\newcommand*{\prob}{\text{P}}

\newlength\barheight 
\newlength\barwidth 

\title{Cognitive Reframing of Negative Thoughts through\\ Human-Language Model Interaction}

\author{Ashish Sharma$^{\spadesuit}$ \: \: \:
    Kevin Rushton$^{\diamondsuit}$ \: \: \:
    Inna Wanyin Lin$^{\spadesuit}$ \: \: \:
    David Wadden$^{\clubsuit}$ \: \: \: \\
   \bf Khendra G. Lucas$^{\diamondsuit}$ \: \: \:
  Adam S. Miner$^{\varheartsuit\heartsuit}$ \: \: \:
  Theresa Nguyen$^{\diamondsuit}$ \: \: \:
  Tim Althoff$^{\spadesuit}$ \\
  $^\spadesuit$Paul G. Allen School of Computer Science \& Engineering, University of Washington \\
  $^\diamondsuit$Mental Health America \quad $^\clubsuit$Allen Institute for Artificial Intelligence \\ 
  $^\varheartsuit$Department of Psychiatry and Behavioral Sciences, Stanford University \\
  $^\heartsuit$Center for Biomedical Informatics Research, Stanford University  \\
  \texttt{\{ashshar,althoff\}@cs.washington.edu}
  }

\begin{document}
\maketitle

\begin{abstract}
A proven therapeutic technique to overcome negative thoughts is to replace them with a more hopeful ``\textit{reframed thought.}'' Although therapy can help people practice and learn this \textit{Cognitive Reframing of Negative Thoughts}, clinician shortages and mental health stigma commonly limit people's access to therapy. In this paper, we conduct a human-centered study of how language models may assist people in reframing negative thoughts. Based on psychology literature, we \textit{define} a framework of seven linguistic attributes that can be used to reframe a thought. We develop automated metrics to \textit{measure} these attributes and validate them with expert judgements from mental health practitioners. We collect a dataset of 600 situations, thoughts and reframes from practitioners and use it to train a retrieval-enhanced in-context learning model that effectively \textit{generates} reframed thoughts and \textit{controls} their linguistic attributes. To investigate what constitutes a ``high-quality'' reframe, we conduct an IRB-approved \textit{randomized field study} on a large mental health website with over 2,000 participants. Amongst other findings, we show that people prefer highly empathic or specific reframes, as opposed to reframes that are overly positive. Our findings provide key implications for the use of LMs to assist people in overcoming negative thoughts.
\end{abstract}
\section{Introduction}
\label{sec:intro}
Negative thoughts are a natural part of human cognition. However, for people experiencing mental health challenges, such thoughts are often entrenched, automatic and emotionally triggering, making it difficult to overcome them in-the-moment \cite{beck1976cognitive}. An evidence-based, well-established therapeutic intervention to overcome negative thoughts is \textit{Cognitive Reframing}, in which a negative thought is replaced with a more hopeful ``\textit{reframed thought}'', which offers an alternative perspective on one's situation \cite{beck1976cognitive}. For example, imagine a person with a situation ``\textit{I'm submitting a research paper to ACL 2023}'' has a thought ``\textit{This paper is going to get rejected}.'' A possible way to reframe this thought is to say ``\textit{This paper has some chance of getting accepted due to its novel methodology and potential impact}.''

Psychotherapy research suggests that for a reframed thought to be effective, it must be (a) relatable to the individual, (b) helpful in overcoming the negative thought and (c) memorable to be accessible the next time a similar thought arises \cite{beck1976cognitive,burns1980feeling}. However, understanding what characterizes a relatable, helpful and memorable reframe is challenging and unknown. Professional therapists can support people in coming up with such highly effective reframed thoughts. However, barriers like clinician shortages, lack of insurance coverage and stigma commonly limit access to therapists \cite{olfson2016building,sickel2014mental}. NLP-based methods that assist individuals in reframing negative thoughts, in-the-moment, may provide scaffolding that is easier to engage with and that could be made widely accessible. 

Prior NLP research has developed methods for a range of text reframing tasks like sentiment and empathy rewriting \cite{reif2022recipe,Sharma2021-rq} and more recently, positive reframing \cite{ziems2022inducing}. However, little is known about how to develop cognitive reframing methods that automatically generate relatable, helpful and memorable reframed thoughts.

\begin{figure*}[t]
\centering
\includegraphics[width=\textwidth]{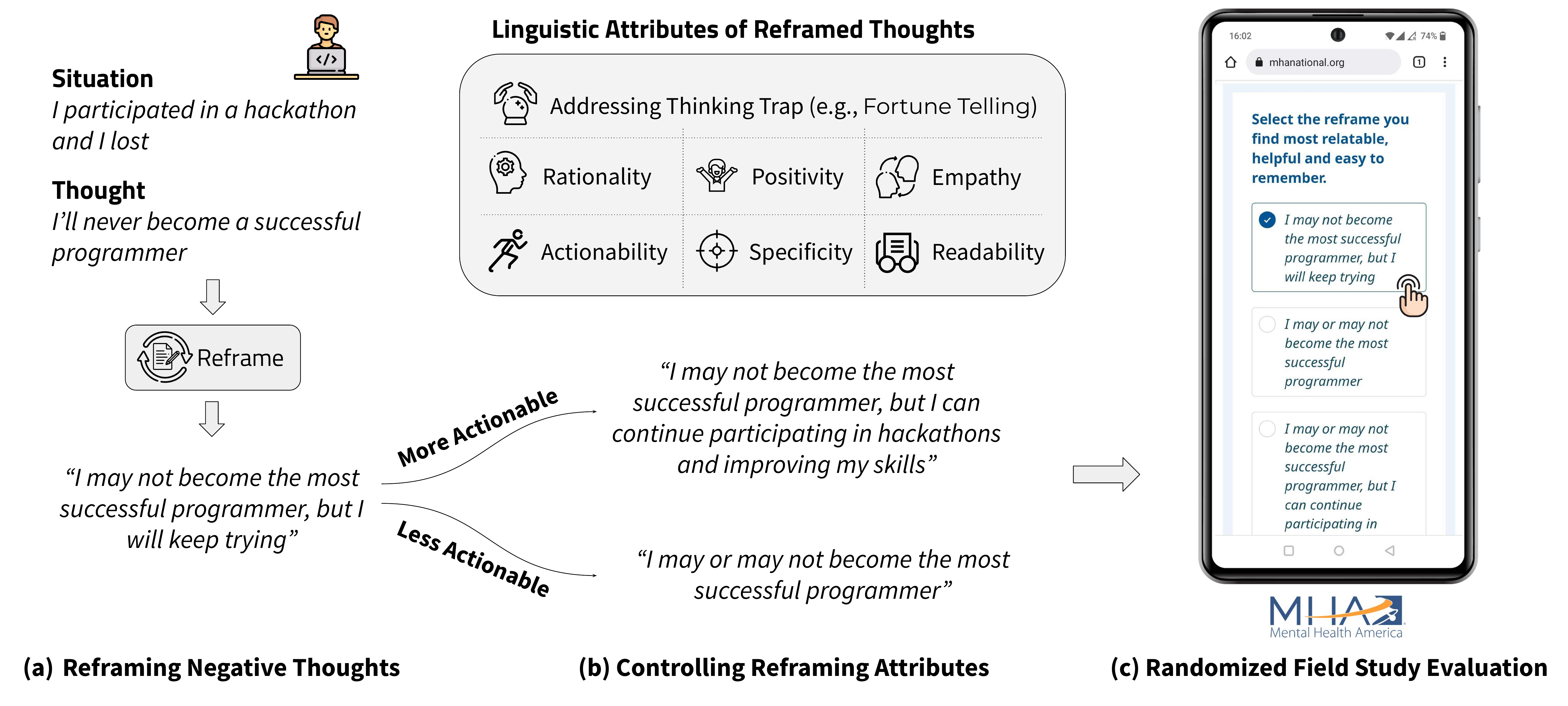}
\vspace{-15pt}
\caption{\textbf{(a)} We consider the task of reframing negative thoughts with different, more hopeful thoughts using LMs; \textbf{(b)} Different perspectives on a situation may result in different reframes. Here, we propose a framework of seven reframing attributes (see gray box). Given a reframed thought, we control each attribute (e.g., \textit{actionability}) to generate reframes that score higher or lower on that attribute (e.g., \textit{more or less actionable}); \textbf{(c)} We deploy this model on Mental Health America, a large U.S. national mental health website (\href{https://bit.ly/changing-thoughts}{bit.ly/changing-thoughts}) and conduct a randomized field study with 2,067 participants. We suggest LM-generated reframes to MHA visitors and assess which reframing attributes are desirable and what constitutes a relatable, helpful and memorable reframe.
}
\vspace{-5pt}
\label{fig:example}
\end{figure*}

In this paper, we conduct a study of how language models can be used to reframe negative thoughts (Figure~\ref{fig:example}). We study ways in which a negative thought can be reframed, how LMs can be utilized to perform this reframing and what types of reframes are preferred by people who experience negative thoughts. 

First, in collaboration with clinical psychologists and mental health professionals, we develop a new conceptual framework for characterizing the ways in which a thought might be reframed. We synthesize the most prominent cognitive reframing processes used in therapy and define seven linguistic attributes of reframed thoughts: whether the reframe \textit{addresses ``thinking traps''} (faulty or distorted patterns of thinking), whether it is \textit{rational}, \textit{positive}, \textit{empathic}, \textit{actionable}, \textit{specific} and \textit{readable}. Building on prior research, we develop automated metrics to measure these attributes and establish construct validity by correlating them with judgements from mental health practitioners.

Next, to develop models for the cognitive reframing task, we collect and share\footnote{Our code and datasets are available at \href{https://github.com/behavioral-data/Cognitive-Reframing-of-Negative-Thoughts}{https://github.com/behavioral-data/Cognitive-Reframing-of-Negative-Thoughts}.} a dataset from mental health practitioners and clinical psychology graduate students. The dataset includes 600 situations and thoughts with expert-suggested reframes as well as annotations of the proposed reframing attributes. Using this dataset, we develop a retrieval-enhanced in-context learning method \cite{brown2020language} to \textit{generate} reframed thoughts and to \textit{control} their linguistic attributes. We show that this method achieves the highest overlap with expert-suggested reframes and the highest relatability and helpfulness ratings based on evaluation from mental health experts, when compared to popular NLP baselines.

We investigate which reframing attributes are desirable and what constitutes a relatable, helpful and memorable reframe. In collaboration (and co-authorship) with mental health experts, and after appropriate ethical review, we deploy a month-long randomized field study on Mental Health America (MHA; a popular website that shares mental health resources and tools online), with 2,067 participants with informed consent. We ask MHA visitors to describe situations and negative thoughts they are experiencing and then suggest LM-generated reframed thoughts with varying linguistic attributes. We find that highly specific and highly empathic reframing is the most preferred and highly specific and actionable reframing is considered the most helpful and memorable. However, we find that reframes that are highly positive are less preferred. These findings provide key implications for cognitive reframing of negative thoughts and for the use of Human-LM interaction in this process.
\section{Problem Definition and Goals}
\label{sec:task}
We work on the task of \textit{Cognitive Reframing}. Given a situation $\mathbf{S_i}$ and a negative thought $\mathbf{T_i}$, the task is to generate a reframed thought $\mathbf{R_i}$.

Psychotherapy literature \cite{beck1976cognitive} highlights three desirable outcomes for a successful reframe: \textbf{(a)} the reframed thought must be \textit{relatable} to the individual, \textbf{(b)} it must \textit{help} them overcome the negative thought and \textbf{(c)} it must be \textit{memorable} the next time a similar negative thinking pattern emerges.

Here, we aim to understand what constitutes successful reframing and how language models can assist people in this process. Towards this goal, we characterize the linguistic attributes of reframed thoughts (\S\ref{sec:framework}), collect a dataset of situations, thoughts and reframes (\S\ref{sec:data}), develop methods to generate reframes and to measure and control their attributes (\S\ref{sec:method}; \S\ref{sec:results}) and investigate which linguistic attributes are related to the reframing outcomes of relatability, helpfulness and memorability (\S\ref{sec:fieldstudy}).
\section{Framework of Linguistic Attributes of Reframed Thoughts}
\label{sec:framework}
We draw from clinical therapy practices and collaborate with mental health experts (some of whom are co-authors) to develop a framework of linguistic attributes of reframed thoughts.
 We illustrate these attributes with the following example for all reframes below -- Situation: ``\textit{I participated in a hackathon and I lost}''; Thought: ``\textit{I'll never become a successful programmer}''. \S\ref{subsec:metrics} will describe automatic metrics to measure these attributes.

\xhdr{Addressing Thinking Traps} Negative thinking often falls into common patterns, called ``\textit{thinking traps}.'' Also called \textit{cognitive distortions}, these include exaggerated and biased patterns of thinking which cause individuals to perceive reality inaccurately \cite{beck1976cognitive,ding2022improving}. Common thinking traps include: assuming what others think (``\textit{Mind reading}''), thinking in extremes (``\textit{All-or-nothing thinking}''), focusing on the worst-case scenario (``\textit{Catastrophizing}''), trying to predict the future (``\textit{Fortune telling}''), etc. See Appendix~\ref{appendix:thinking-traps} for the full list.

A reframe may or may not directly address one or more of the thought's thinking traps. A reframe like ``\textit{I don't know what the future will bring}'' directly addresses the thinking trap ``\textit{Fortune telling},'' whereas a reframe like ``\textit{I will surely become a successful programmer}'' does not address this thinking trap but rather continues to express it.

\xhdr{Rationality} Another strategy to reframe a thought is to reflect on evidence for and against it and reason about what these evidence imply \cite{beck1976cognitive}. For example, losing the hackathon is one evidence of having the thought ``\textit{I'll never become a successful programmer}.'' However, the evidence against this thought could be that winning or losing a single hackathon does not make someone a failure, which may lead to a reframe ``\textit{Just losing one hackathon doesn't define my failure}.'' A rational reframe is guided by such strong evidence whereas an irrational reframe is based on unrealistic assumptions.

\xhdr{Positivity} A reframe of a negative thought tries to emphasize the positive perspectives on the situation but different reframes may have different levels of positivity. 
An overly positive reframe like ``\textit{I'm going to win every hackathon from now on}'' exaggerates the positive perspectives, which is likely to set the person up for disappointment rather than success \cite{dember1980happiness}. On the other hand, a balanced response like ``\textit{I may or may not succeed, but I'll keep trying}'' considers both positive and negative perspectives of the situation.

\xhdr{Empathy} It can be helpful to acknowledge the feelings caused by negative thoughts \cite{allen2010self,elliott2011empathy}. A reframe may express empathy or self-compassion by validating how one is feeling. E.g., ``\textit{It is okay to feel disappointed for not winning the hackathon}.''

\xhdr{Actionability} To encourage pleasant emotions, one commonly used therapeutic approach is Behavioral Activation \cite{dimidjian2011origins,burkhardt2021behavioral}. This involves engaging in behaviors or actions that may help in overcoming negative thoughts. A reframe may suggest specific actions (e.g., ``\textit{I can continue to practice and participate in hackathons}''), may not suggest specific actions but be actionable (e.g., ``\textit{I may not be very successful, but I can keep trying}'') or may not be actionable at all (e.g., ``\textit{I may or may not become a successful programmer}'').

\xhdr{Specificity} A reframe may specifically address the situation and the thought (e.g., ``\textit{One hackathon doesn't define my failure as a programmer}'') or may be generic enough to be applicable to a wide range of negative situations and thoughts (e.g., ``\textit{I'm going to succeed}''). While a specific reframe may be more helpful in-the-moment, a generic reframe could be effective for recurring thoughts, which are frequently a result of the ``core'' beliefs that a person holds \cite{beck2005cognitive,david2009rational}.

\xhdr{Readability} The linguistic reasoning capabilities of individuals may be different (e.g., across age groups or education levels) \cite{kaplan1995cognitive}. Accordingly, a reframe may either be simple or complex to read (e.g., ``\textit{I'll do well in the future}'' vs. ``\textit{I'm resolute in my ambition to succeed}'').

\section{Data Collection}
\label{sec:data}
To facilitate computational methods for cognitive reframing, we collect a dataset of reframed thoughts, annotated with their linguistic attributes.

\subsection{Curated Situations \& Negative Thoughts} 
\label{subsec:datasource}
We start by curating data sources for situations and negative thoughts.

\xhdr{Thought Records Dataset \cite{burger2021natural}} This dataset contains hypothetical and real-world situations, thoughts and emotional processes reported by crowdworkers on Amazon Mechanical Turk. We manually curate 180 pairs of diverse situations with negative thoughts from this dataset.

\xhdr{Mental Health America (MHA)} Situations and thoughts from crowdworkers may not reflect the broad range of mental health challenges that people face in real-life. To incorporate more real-world situations and thoughts, we ran a survey on the MHA website (\href{https://screening.mhanational.org/}{screening.mhanational.org}). MHA visitors (who typically use the website for screening of mental illnesses) were asked to describe any negative thoughts and the associated situations they were struggling with. We manually curate 120 pairs of self-reported situations and thoughts to ensure broad coverage of relevant topics based on high diversity and manual filtering.

\subsection{Annotation Task and Procedure}
\label{subsec:annotation_procedure}
Reframing negative thoughts is a cognitively difficult process that requires practice and training, making crowdwork data collection approaches challenging. To ensure high-quality reframes and annotations, we recruit 15 current mental health practitioners and clinical psychology graduate students with significant practical experience in cognitive reframing.\footnote{For recruitment, we advertised our study through university mailing lists and newsletter of a mental health organization. Recruited experts were paid @ 37.5 USD / hr.} For each (situation, thought) pair in our data source (\S\ref{subsec:datasource}), we ask them to (1) write two different reframed thoughts, (2) annotate the thinking traps addressed by each reframed thought and (3) compare the two reframes and choose the one that is more rational, more positive, more actionable, more empathic, more specific and more readable. In total, we collect 600 reframed thoughts with annotations on their linguistic attributes. We share this dataset publicly at \href{https://github.com/behavioral-data/Cognitive-Reframing-of-Negative-Thoughts}{https://github.com/behavioral-data/Cognitive-Reframing-of-Negative-Thoughts}. 

\subsection{Ethics and Safety} 
Our data collection and randomized field study (\S\ref{sec:fieldstudy}) were designed and conducted after review of potential benefits and risks to participants in consultation and collaboration with mental health experts. Both studies were approved by the University of Washington's Institutional Review Board and informed participants about study purpose, risks and data collection. All participants were 18 or older, provided informed consent and were given access to a crisis hotline. We do not assess any clinical outcomes. See \S\ref{sec:ethics} for an extended discussion of ethical and safety considerations.
\section{Method}
\label{sec:method}

We design automated metrics for linguistic attributes (\S\ref{subsec:metrics}), develop methods to generate reframed thoughts (\S\ref{subsec:reframe-generation}) and to control their attributes (\S\ref{subsec:control-reframe}).

\subsection{Measuring Reframing Attributes}
\label{subsec:metrics}
\xhdr{Addressing Thinking Traps} Given a situation $\mathbf{S_i}$, a negative thought $\mathbf{T_i}$ and a reframed thought $\mathbf{R_i}$, our goal is to identify the set of thinking traps addressed by the reframed thought. We approach this as a multi-label classification task, and fine-tune a GPT-3 model\footnote{We use \texttt{text-davinci-003} as our GPT-3 model for all experiments in this paper.} on the expert-annotated thinking trap labels collected in \S\ref{subsec:annotation_procedure}.

\xhdr{Rationality} Rationality is the quality of being guided by reasons \cite{damielson2004oxford}. Here, we operationalize rationality of a reframed 
thought $\mathbf{R_i}$ as its \textit{reasoning strength} and ask the following two questions: (1) What might be the reasoning behind $\mathbf{R_i}$?; (2) Are the reasons sound? To understand the reasoning behind $\mathbf{R_i}$, we develop an \textit{abductive explanation} based method \cite{peirce1974collected,bhagavatula2020abductive,jung-etal-2022-maieutic}. For a given ($\mathbf{S_i},\mathbf{T_i}$), we use a language model to generate (a) the most plausible explanations that \textit{support} $\mathbf{R_i}$ and (b) the most plausible explanations that \textit{refute} it. Moreover, to check if the explanations are sound, we recursively generate explanations behind the explanations to test their reasoning strength (Appendix~\ref{appendix:rationality}). Let $sup(\cdot)$ be a generator function that generates explanation \textit{supporting} a reframe and let $ref(\cdot)$ be a generator function that generates explanation \textit{refuting} a reframe. Then, we recursively define reasoning strength $RS(\mathbf{S_i}, \mathbf{T_i}, \mathbf{R_i})$ as

\vspace{-10pt}
\begin{small}
\begin{align}
   & \left( \prob(\mathbf{R_{i}}=\texttt{sound} | \mathbf{S_i}, \mathbf{T_i}) \times  \displaystyle\mathbb{E}_{r \sim sup(\cdot)} \left[ RS\left(\mathbf{S_i}, \mathbf{T_i}, r \right) \right] \right) \nonumber \\
    -& \left(\prob(\mathbf{R_{i}} = \texttt{flawed} | \mathbf{S_i}, \mathbf{T_i}) \times \displaystyle\mathop{}{\mathbb{E}}_{r \sim ref(\cdot)} \left[ RS\left(\mathbf{S_i}, \mathbf{T_i}, r \right) \right] \right) \nonumber
\end{align}
\end{small}
\noindent To design the explanation generator functions, $sup(\cdot)$ and $ref(\cdot)$, we leverage in-context learning \cite{brown2020language}. In collaboration with mental health experts, we design 10 demonstration examples of situations, thoughts and reframed thoughts with explanations that support (``\textit{This reframed thought is sound because...}'') and refute (``\textit{This reframed thought is flawed because...}'') a particular reframe. We use these examples to prompt GPT-3. Moreover, to estimate the probabilities $\prob(\mathbf{R_{i}} = \texttt{sound})$ and $\prob(\mathbf{R_{i}} = \texttt{flawed})$, we use the token probability of generating ``\textit{sound}'' and ``\textit{flawed}'' respectively, given $\mathbf{S_i}, \mathbf{T_i}, \mathbf{R_{i}}$ and the text ``\textit{This reframed thought is}'' as input to GPT-3.\footnote{We experimented with different alternatives for ``\textit{sound}'' and ``\textit{flawed}'' and observed similar results.}

\xhdr{Positivity} To measure the positivity of the generated reframed thought, we use a RoBERTa-based sentiment classifier fine-tuned on the TweetEval benchmark \cite{barbieri-etal-2020-tweeteval}.


\xhdr{Empathy} To measure empathy, we build upon the empathy classification model presented in \citet{sharma2020computational}. This RoBERTa-based model leverages a theoretically-grounded framework of empathy consisting of three empathy communication mechanisms (emotional reactions, interpretations and explorations) and predicts empathy levels in mental health conversations on a scale from 0 to 6. Here, we further fine-tune this model on the domain of reframed thoughts through a manually labeled dataset of 300 reframed thoughts with empathy labels (labeled by one author with expertise in empathy in mental health context).

\xhdr{Actionability}  To measure actionability, we hypothesize that an actionable reframe is one that either (1) \textit{suggests a concrete action} or (2) does not suggest a concrete action but is \textit{easy to act upon}.

We cast action concreteness as a binary classification task: given reframe $\mathbf{R_i}$, predict $contains\_action(\mathbf{R_i}) \in \{0, 1\}$. We make few-shot predictions by prompting GPT-3 with 10 examples of reframed thoughts paired with actionability ratings from \S\ref{subsec:annotation_procedure} (details in Appendix \ref{appendix:linguistic_attributes}).

To determine the ease with which $\mathbf{R_i}$ can be acted upon, we examine the next set of actions entailed by $\mathbf{R_i}$. Our hypothesis is that a \textit{diverse} next action set may indicate ambiguity which might be less actionable, whereas a \textit{coherent} next action set may indicate clarity which might be more actionable. Here, we instruct GPT-3 to generate $k=5$ next action candidates given a reframed thought (instruction prompting; zero-shot). We compute the next action coherence --- denoted $next\_action\_coherence(\mathbf{R_i})$ --- by embedding each of the $k$ action candidates using RoBERTa \cite{liu2019roberta} and computing the average pairwise cosine similarity. Higher similarity indicates greater coherence among the possible next actions.
Our overall actionability measurement is defined as $contains\_action(\mathbf{R_i}) + next\_action\_coherence(\mathbf{R_i})$.

\xhdr{Specificity} Following prior work~\cite{xu2018better, Sharma2021-rq}, we measure specificity using sentence embedding similarity between the reframed thought $\mathbf{R_i}$ and the concatenation of the situation $\mathbf{S_i}$ and the thought $\mathbf{T_i}$ (using RoBERTa embeddings~\cite{liu2019roberta}).

\xhdr{Readability} We employ the commonly used Coleman-Liau Index (CLI) metric \cite{coleman1975computer} which assesses readability based on the character and word structure within a sentence. The Coleman-Liau Index is calculated as $0.0588L - 0.296S -15.8$, where $L$: average number of letters per 100 words; $S$ is the average number of sentences per 100 words.

\subsection{Reframe Generation}
\label{subsec:reframe-generation}
In-context learning methods can learn to generalize NLP tasks from a handful of examples (\textit{few-shot learning}) or from hand-written instructions alone (\textit{instruction prompting}) \cite{brown2020language}. However, through a qualitative analysis of 100 manually written situations and thoughts, we found that a simple in-context learning method with a fixed set of examples often failed to appropriately reframe situations and thoughts for which no relevant in-context examples were provided (e.g., someone with anxiety having ``\textit{racing thoughts}'').

To appropriately reframe thoughts related to a range of situations and thoughts, we develop a retrieval-based in-context learning method \cite{liu2022makes}. For each situation $\mathbf{S_i}$ and negative thought $\mathbf{T_i}$, we retrieve $k$-similar examples from our dataset (\S\ref{sec:data}). We first encode situations and thoughts using RoBERTa embeddings. Then, we retrieve $k$ examples, $\left\{(s_1, t_1), ..., (s_k, t_k)\right\}$, from our dataset based on the top-$k$ values of  $cosine\_sim(s, \mathbf{S_i}) * cosine\_sim(t, \mathbf{T_i})$. We choose $k=5$ (Appendix~\ref{appendix:hyperparameters}).

\subsection{Controlling Linguistic Attributes of Generated Reframes}
\label{subsec:control-reframe}

\begin{table}
\centering
\resizebox{0.9\columnwidth}{!}{
\def\arraystretch{1.2}
\begin{tabular}{lc}
\toprule
\textbf{Attribute} & \textbf{Pearson Correlation} \\
\toprule
Addressing Thinking Traps & 0.680*** \\
Rationality & 0.448** \\
Positivity & 0.550*** \\
Empathy & 0.575*** \\
Actionability & 0.647*** \\
Specificity & 0.427** \\
Readability & 0.331* \\
\bottomrule
\end{tabular}
}
\caption{Correlation of our proposed attribute measures by with human judgments from mental health experts. *:$p<0.05$; **:$p<0.001$; ***:$p<10^{-5}$.}
\label{tab:attribute-correlation}
\vspace{-5pt}
\end{table}


While our proposed method allows us to generate a single reframe, it does not directly give us control over its linguistic attributes beyond mimicking the retrieved examples (\S\ref{sec:framework}). Here, we intend to vary the linguistic attributes of the reframes. 

A reframed thought may or may not address a thinking trap in the original thought $\mathbf{T_i}$. 
Here, we generate two reframes $\mathbf{R_i}^{(tt,\text{Y})}$ and $\mathbf{R_i}^{(tt,\text{N})}$, one that addresses the thinking traps in $\mathbf{T_i}$ and another that does not address it.\footnote{If a thought exhibits multiple thinking traps, we check if the reframe addresses at least one of them.} We extract two separate sets of in-context examples from our dataset -- those that address at least one thinking trap and those that do not (as collected in \S\ref{sec:data}). We use those examples to prompt GPT-3 to generate $\mathbf{R_i}^{(tt,\text{Y})}$ and $\mathbf{R_i}^{(tt,\text{N})}$.



Moreover, a reframed thought may have high or low rationality, positivity, empathy, actionability, specificity and readability values. For these six attributes, given a reframe $\mathbf{R_i}$ and a linguistic attribute $a$, we generate two \textit{additional} reframes $\mathbf{R_i}^{(a, \text{H})}$ and $\mathbf{R_i}^{(a, \text{L})}$, one that scores higher on attribute $a$ and another that scores lower on it (e.g., higher or lower actionability). 
To accomplish this, recall that each (situation, thought) pair from \S \ref{subsec:annotation_procedure} is annotated with two reframes and that the reframes are compared along each linguistic attribute. For a human-annotated instance $j$, let $\mathbf{R_j}^{*(a, \text{H})}$ and $\mathbf{R_j}^{*(a, \text{L})}$ be the reframes judged to be high and low on attribute $a$, respectively. To generate $\mathbf{R_i}^{(a, \text{H})}$ from $\mathbf{R_i}$, we prompt GPT-3 with in-context examples $\{ \mathbf{R_j}^{*(a, \textrm{L})} \rightarrow  \mathbf{R_j}^{*(a, \textrm{H})} \}_{j=1}^k$, using $k=5$. Similarly, to generate $\mathbf{R_i}^{(a, \text{L})}$ from $\mathbf{R_i}$, we prompt GPT-3 with examples $\{\mathbf{R_j}^{*(a, \text{H})} \rightarrow  \mathbf{R_j}^{*(a, \text{L})} \}_{j=1}^k$.

\section{Experiments and Results}
\label{sec:results}

We assess the construct validity of proposed linguistic attributes (\S\ref{subsec:construct-val}) and evaluate the performance of the reframe generation model (\S\ref{subsec:generation-eval}).

\subsection{Construct Validity of Linguistic Attributes}
\label{subsec:construct-val}
We validate our proposed linguistic attribute measures by correlating them with the human judgments of mental health experts, as obtained in \S\ref{subsec:annotation_procedure}. We find a strong Pearson correlation for addressing thinking traps (0.680***) and actionability (0.647***), a moderate correlation for rationality (0.448**), positivity (0.550***), empathy (0.575***) and specificity (0.427**), and a weak correlation for readability (0.331*) (Table~\ref{tab:attribute-correlation}).\footnote{*:$p<0.05$;**:$p<0.001$;***:$p<10^{-5}$}

\begin{table}
\small
\centering
\def\arraystretch{1.15}
\resizebox{1.03\columnwidth}{!}{
\begin{tabular}
{l|cccc|cc}
\toprule
\multirow{2}{*}{\textbf{Model}} & \multicolumn{4}{c|}{\multirow{1}{*}{\parbox{2 cm}{\centering \textbf{Automatic}}}} & \multicolumn{2}{c}{\multirow{1}{*}{\parbox{1.3 cm}{\centering \textbf{Human}}}}  \\
& BLEU & R-1 & R-L & BScore & Rel. & Help. \\
\midrule
Retrieval Only & 21.6 & 18.8 & 14.2 & 86.7 & 2.58 & 3.14 \\
\midrule
Pos. Reframing & 24.4 & 23.6 & 17.6 & 87.6 & 2.67 & 2.40 \\
\midrule
DialoGPT & 22.5 & 17.4  & 13.5 & 86.3 & 2.49 & 3.21 \\
T5 & 24.9 & 23.4 & 17.8 & 87.2 & 2.51 & 3.30 \\
\midrule
GPT-3 Only & 25.0 & 23.9 & 18.0 & 88.3 & 2.97 & 3.98 \\
Our Model & \textbf{27.8} & \textbf{26.0} & \textbf{19.9} & \textbf{88.6} & \textbf{3.10} & \textbf{4.11} \\
\bottomrule
\end{tabular}
}
\caption{Automatic and Human Evaluation Results. R-1: ROUGE-1; R-L: ROUGE-L; BScore: BertScore; Rel.: Relatability; Help.: Helpfulness.} 
\label{tab:model-performance}
\vspace{-5pt}
\end{table}

\begin{figure*}[t]
\centering
\includegraphics[width=0.9\textwidth]{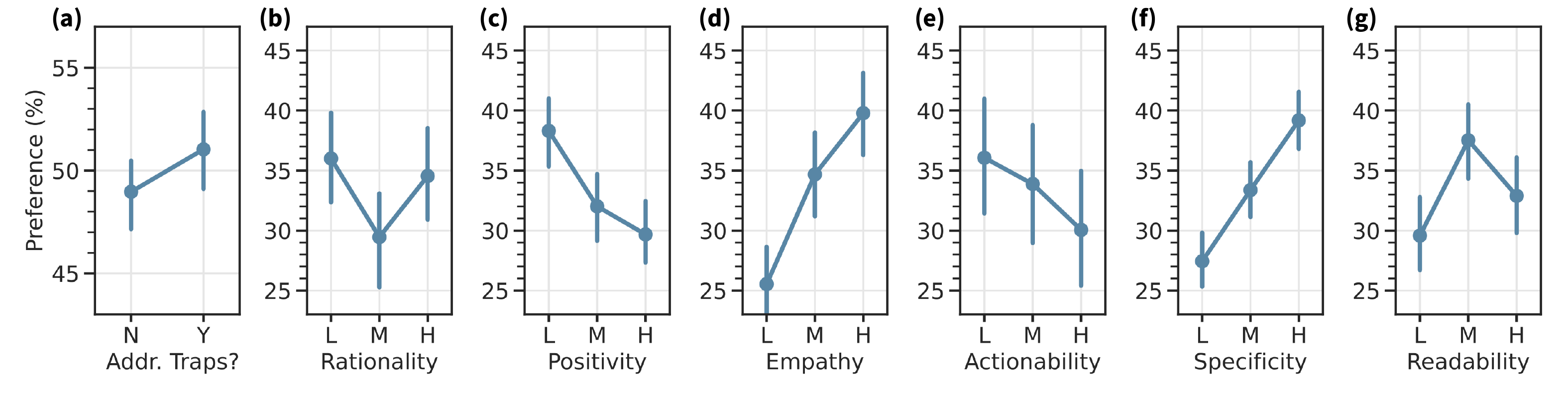}
\vspace{-5pt}
\caption{\textbf{Which linguistic attributes of reframed thoughts do people prefer?} For a given situation and thought from MHA visitors, we show them multiple LM-generated reframes with variance across a randomly selected attribute (e.g., low, medium and high actionability). We find that highly empathic and highly specific reframings are more preferred. On the other hand,  reframes with high positivity are less preferred. N: No; Y: Yes; L: Low; M: Medium; H: High. Error bars in any figure represent 95\% bootstrapped confidence intervals.}
\label{fig:preference}
\end{figure*}

\subsection{Reframe Generation Performance}
\label{subsec:generation-eval}
We use both automatic and human evaluation to assess the performance of our proposed reframe generation model as developed in \S\ref{subsec:reframe-generation}.

\xhdr{Experimental Setup} We use top-$p$ sampling with $p=0.6$ for text generation \cite{holtzman2019curious}. We split the 600 expert-annotated examples (\S\ref{sec:data}) into train and test using a 70:30 split. 

\xhdr{Baselines} \textbf{(1)} \textit{Retrieval Only} --  For a test input, we retrieve the training set example with the highest cosine similarity based on RoBERTa embeddings; \textbf{(2)} \textit{Positive Reframing} -- We reuse the BART-based positive reframing model from ~\citet{ziems2022inducing}; \textbf{(3)} \textit{DialoGPT} -- GPT-2 adapted to dialogue \cite{zhang2019dialogpt}; \textbf{(4)} \textit{T5} -- Text-to-text transfer LM~\cite{raffel2020exploring};\footnote{Training DialoGPT and T5 on 600 samples only may be challenging. Here, we use an \textit{overgeneration} strategy -- Starting from our collected data samples, we utilize the pattern replication capabilities of GPT-3 to generate 10,000 more examples, similar to \citet{liu2022wanli}.} \textbf{(5)} \textit{GPT-3 Only} -- We randomly retrieve 5 examples from our training set and use them to prompt GPT-3.


\xhdr{Automatic Evaluation} We examine the semantic similarity between the model outputs and the ground truth reframings in the above-created test split. We use BLEU \cite{papineni2002bleu}, ROUGE-1, ROUGE-L \cite{lin2004rouge} and the
BERTScore \cite{zhang2019bertscore} metrics. We find that our proposed model has an 11.2\% higher BLEU score and 9.7\% higher ROUGE scores than the next best-performing baselines -- GPT-3 Only and Positive Reframing (Table~\ref{tab:model-performance}).

\xhdr{Human Evaluation} We assess the two key reframing outcome metrics of \textit{relatability} (how relatable would a reframed thought be) and \textit{helpfulness} (how helpful would a reframed thought be in overcoming negative thoughts). We recruit three mental health practitioners. We ask them to rate the models' outputs on test set examples based on their reliability and helpfulness on a 1 to 5 scale. We find that our proposed model achieves the highest relatability and helpfulness ratings (Table~\ref{tab:model-performance}). Surprisingly, the Positive Reframing method showed the least helpfulness and low relatability, indicating that just reframing negative thoughts based on positivity may not be highly relatable and helpful.

\section{Randomized Field Study on a Large Mental Health Platform}
\label{sec:fieldstudy}

Next, we deploy our model on a large mental health platform (\S\ref{subsec:fieldstudy}) and study what types of reframes do people prefer (\S\ref{subsec:select}) and what characterizes relatable, helpful and memorable reframes (\S\ref{subsec:outcomes}).

\subsection{Model Deployment}
\label{subsec:fieldstudy}

We try to understand how our proposed cognitive reframing model may assist people who experience negative thoughts. After careful assessment of ethical and safety considerations, active collaboration with mental health experts and clinical psychologists (some of whom are co-authors) and IRB approval, we deploy our model on Mental Health America (MHA), a large mental health website that provides mental resources and tools to millions of users (\href{https://bit.ly/changing-thoughts}{bit.ly/changing-thoughts}). We conduct a month-long randomized field study with 2,067 MHA visitors as participants. After choosing to use our model and after consenting to participate, MHA visitors described their situation and the thoughts they were struggling with. Next, they were shown multiple model-generated reframed thoughts in random order, asked to select the reframed thought they find most relatable, helpful and memorable and finally evaluate the selected reframed thought based on relatability, helpfulness and memorability (See Appendix~\ref{appendix:tool-interface}).

\begin{figure*}[t]
\centering
\includegraphics[width=0.9\textwidth]{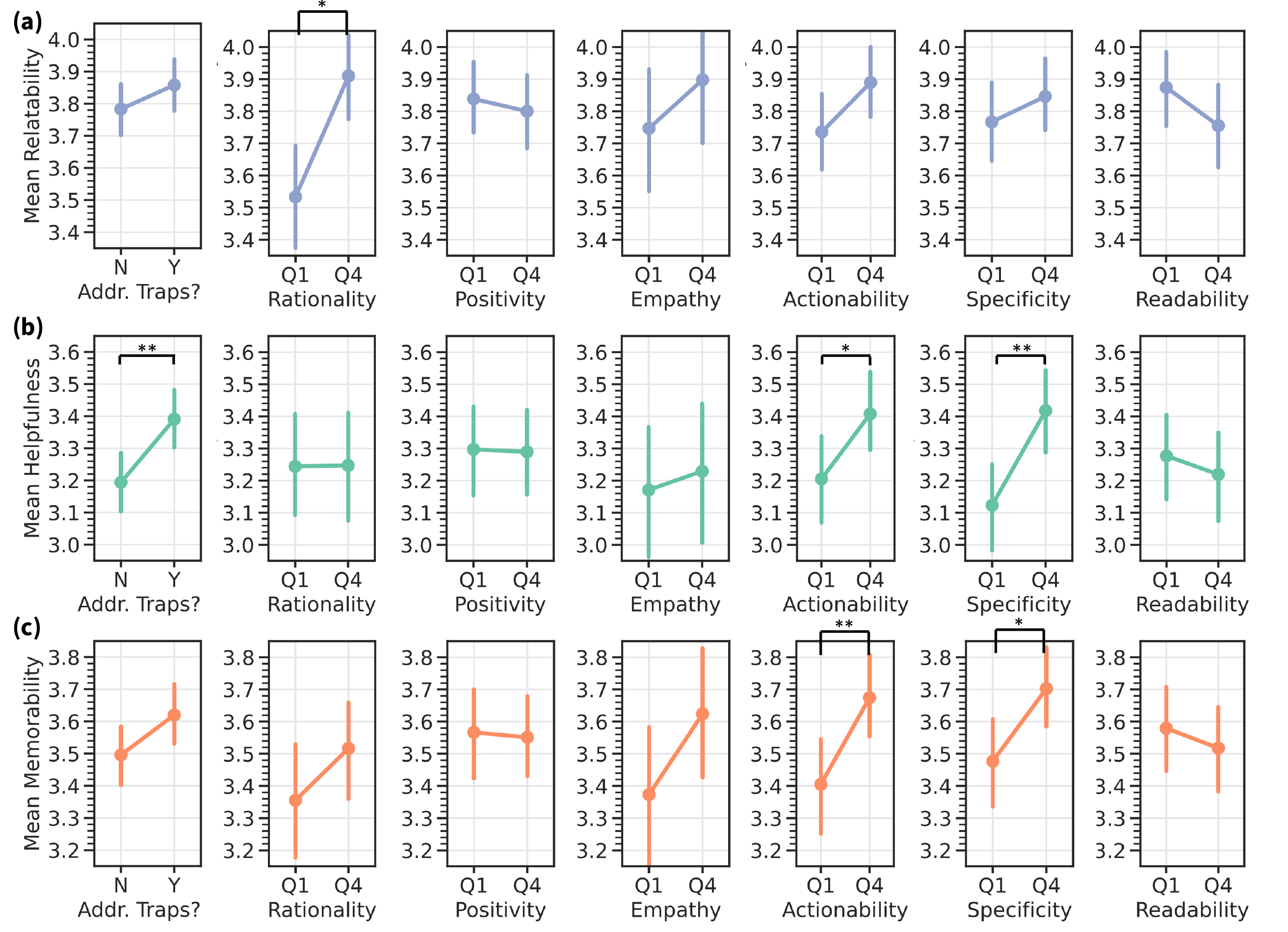}
\vspace{-5pt}
\caption{\textbf{Which linguistic attributes are associated with desired cognitive reframing outcomes?} For a given situation and thought, we show one LM-generated reframe to MHA participants and ask them to rate it on relatability, helpfulness and memorability on a 1 to 5 scale. For each linguistic attribute, we compare the first (Q1) and the fourth quartile (Q4). We find that \textbf{(a)} reframes that have higher rationality are more relatable; \textbf{(b)} reframes that address thinking traps, have higher actionability or higher specificity are more helpful; \textbf{(c)} reframes that have higher actionability or higher specificity are more memorable. *:$p<0.05$;**:$p<0.01$. }
\label{fig:outcome}
\vspace{-5pt}
\end{figure*}

\subsection{What types of reframed thoughts do people prefer?}
\label{subsec:select}

To understand which reframing attributes people prefer, we suggest multiple LM-generated reframes which vary across our attribute values. Given a situation and thought, we start by generating one reframed thought using our model. Next, we randomly select an attribute (e.g., actionability) and vary the first reframe based on it (e.g., to generate two additional reframes with higher or lower actionability) using our proposed controllable text generation method (\S\ref{subsec:control-reframe}). Figure~\ref{fig:preference} reveals key differences between the linguistic attributes of reframes that people select and prefer:


\xhdr{(1) Highly empathic and specific reframings are preferred more} We find that highly empathic reframes are preferred 55.7\% more frequently than reframes with lower empathy (39.7\% vs. 25.5\%; $p<10^{-5}$); highly specific reframes are preferred 43.1\% more frequently than reframes with lower specificity (39.2\% vs. 27.4\%; $p<10^{-5}$). Prior work has shown the importance of empathy and less ``templated'' responses in mental health support conversations \cite{sharma2020computational,althoff2016large}. Here, we show that empathy and specificity of LM-generated reframes may support people in reframing negative thoughts. 

\xhdr{(2) Overly positive reframes are preferred less} On the other hand, reframes with high positivity are preferred 22.7\% less frequently than reframes with lower positivity (29.6\% vs. 38.3\%; $p < 10^{-5}$). This may be because adopting an overly positive reframed thought may be challenging for individuals who are already experiencing emotionally triggering negative thoughts \cite{dember1980happiness}. 

Participants also prefer medium-readability reframes over very simple or very complex reframes, perhaps because their language is balanced for a wider audience. 


\subsection{How do the linguistic attributes of reframed thoughts relate to the desired outcomes of cognitive reframing?}
\label{subsec:outcomes}

We assess what characterizes a reframe that is relatable, helpful and memorable. We show a \textit{single} model-generated reframe to the participants and ask them to rate it on a 5-point Likert scale \cite{likert1932technique} with regards to the three outcome measures (1: Strongly Disagree; 5: Strongly Agree). We do not provide participants in this experiment with a choice of multiple reframes to avoid any selection effects (\S\ref{subsec:select}).
Figure~\ref{fig:outcome} offers key insights on which attributes of reframed thoughts are related to different desired outcomes:

\xhdr{(1) Reframes that are more rational are more \textit{relatable}} We find that reframes that have higher rationality are 10.8\% more relatable than lower rationality reframes (3.91 vs. 3.53; $p<0.05$). This may be because higher rationality reframes, by definition, are more likely to be based on reasons and are less likely to make unrealistic assumptions, making them easier to relate to.

\xhdr{(2) Reframes that address thinking traps and are more actionable and specific are more \textit{helpful}} Reframes that address thinking traps are 6.3\% more helpful than reframes that do not address them (3.39 vs. 3.19; $p<0.01$). Such reframes specifically challenge the cognitive biases in thinking patterns (e.g., ``\textit{Fortune Telling}''; Appendix~\ref{appendix:thinking-traps}), which has shown to be more effective in dealing with negative thoughts in psychotherapy research \cite{beck1976cognitive,burns1980feeling}. Moreover, we find that reframes with higher actionability are 6.6\% more helpful than lower actionability reframes (3.41 vs. 3.20; $p<0.05$) and reframes with higher specificity are 9.6\% more helpful than lower specificity reframes (3.42 vs. 3.12; $p<0.01$). 

\xhdr{(3) Reframes that are more actionable and more specific are more \textit{memorable}} We find that reframes with higher actionability are 7.9\% more memorable than lower actionability reframes (3.67 vs. 3.40; $p < 0.01$) and reframes with higher specificity are 6.3\% more memorable than lower specificity reframes (3.70 vs. 3.48; $p < 0.05$). 

\vspace{-2.5pt}
\section{Related Work}
\label{sec:related}
\vspace{-2.5pt}
Several Human-LM interaction tools for mental health assist support providers, e.g., clinicians \cite{tanana2019development,shen2020counseling} or peers \cite{Sharma2023}. Our work provides insights on how Human-LM interaction may directly support people struggling with mental health challenges through cognitive reframing. Computational work on cognitive reframing has relied on small-scale crowdsourcing studies \cite{smith2021effective,morris2015efficacy}. Our work develops scalable methods for cognitive reframing and conducts a randomized field study on a large mental health platform. Prior text reframing research has developed methods for related tasks including style, sentiment, politeness and empathy transfer \cite{reif2022recipe,madaan2020politeness,Sharma2021-rq} as well as positive reframing \cite{ziems2022inducing}. Our work develops text-reframing methods for cognitive reframing and demonstrates that linguistic attributes of addressing thinking traps, rationality, actionability, specificity and readability are critical to high-quality reframes. More broadly, our work relates to the growing body of research in NLP for mental health and psychological well-being \cite{althoff2016large,sharma2018mental,gaur2019knowledge,lee2019identifying,miner2019key,pendse2019cross,perez2019makes,pruksachatkun2019moments,yang2019channel,zhang2019finding,jaidka2020beyond,saha2020causal,sharma2020engagement,sharma2020computational,wadden2020effect,welch2020expressive,zhang2020balancing,lahnala2021exploring,lin-etal-2022-gendered,naseem2022early,perez2022pair,shah2022modeling,shen2022knowledge,stewartexpressive}.
\vspace{-2.5pt}
\section{Conclusion}
\label{sec:discussion}
\vspace{-2.5pt}
In this paper, we conducted a study of how Human-Language Model Interaction may support humans in the cognitive reframing of negative thoughts. We define a framework of seven linguistic attributes of cognitive reframing, develop automatic metrics to measure these attributes and validate their measurements with mental health experts. We collect and share a dataset of 600 situations, thoughts and reframes from mental health experts and use it to train a retrieval-enhanced in-context learning model based on GPT-3. We deploy this model on the Mental Health America website and conduct a randomized field study with 2,067 participants. We find that people struggling with negative thoughts prefer reframes that are highly empathic or specific, but do not prefer reframes that are highly positive.
\section{Ethics Statement}
\label{sec:ethics}
Intervention in high-risk settings such as mental health necessitates ethical considerations related to safety, privacy and bias. There is a possibility that, in attempting to assist, AI may have the opposite effect on people struggling with mental health challenges. Here, in active collaboration and consultation with mental health professionals and clinical psychologists, we took several measures to minimize these risks.   

\xhdr{IRB Approval} We obtained approval from the University of Washington's Institutional Review Board for both our data collection (IRB ID STUDY00015882) as well as the randomized field study (IRB ID STUDY00016783). Our organization requires all research personnel who conduct human subjects research to complete human subjects protection training using the online CITI course. The graduate students conducting these studies were certified by our IRB.

\xhdr{Informed Consent from Participants} We obtained informed consent from all participants in our randomized field study (Appendix~\ref{appendix:consent}). All participants were 18 years of age and older. Participants were informed that they will be interacting with an AI-based model that automatically generates reframed thoughts and is not monitored by a human. Also, they were informed about the possibility that some of the generated content may be upsetting or disturbing. 

\xhdr{Crisis Resources} We made it very explicit that the model should not be used as a ``cry for help'' outlet and should not be used in cases of suicidal ideation and self-harm. Also, we provided two crisis resources -- Crisis Text Line (\href{https://www.crisistextline.org/}{crisistextline.org}) and 988 Suicide and Crisis Lifeline (\href{https://988lifeline.org/}{988lifeline.org}) -- to our participants at the start of the study.

\xhdr{Safety Measures} To minimize harmful LM-generated reframings, we filtered out any response that contained suicidal ideation or self-harm-related words or phrases. For this, we created a list of 50 regular expressions (e.g., to identify phrases like ``\textit{feeling suicidal}'', ``\textit{wish to die}'', ``\textit{harm myself}'') using suicidal risk assessment lexicons such as \citet{gaur2019knowledge}. An LM-generated response that matched any of the regular expressions was filtered out and not shown to the participants. Also, participants were given an option to flag inappropriate reframing suggestions through a ``Flag inappropriate'' button (Appendix~\ref{appendix:flagged-reframes}).

\xhdr{Privacy} We did not collect any privately identifiable information in our randomized field study and removed any user identifiers before conducting our data analysis. All research data was stored within a separate secure computing environment and only trained research personnel were provided access to data. The situations and thoughts collected in \S\ref{subsec:datasource} went through an anonymization process, where we manually removed any user identifiers and replaced any specific identifiable information including locations, names, etc. with their more general version, following \citet{matthews2017stories}.

\section{Limitations}
\label{sec:limitations}

We conducted our randomized field study on a single platform (Mental Health America) and in a single language (English). However, MHA is a particularly popular source for mental health resources with over five million yearly visitors.


In addition, we note that a range of socio-cultural factors might influence how negative thoughts should be reframed and how LMs assisting this process should be developed. Conducting studies on specific communities, including underrepresented communities and minorities, was beyond the scope of this research. Ensuring equitable access of these tools and adapting them to various socio-cultural contexts requires further investigation.

Not all cognitive reframing implementations elicit situations, but we believed it was essential for making the reframe personally relatable. In the future, when an individual uses the system for multiple situations and thoughts, it would be interesting to study how their context can be learned more effectively over time. Due to privacy concerns, we presently do not gather information to link multiple sessions. However, with appropriate ethical considerations and user consent, this approach may be beneficial.


Our focus in this paper was primarily on creating an intervention that is effective in-the-moment. This was motivated by recent clinical psychology research that suggests that such single-session, in-the-moment interventions can lead to significant positive long-term mental health outcomes~\cite{schleider2022randomized}. To integrate a partial longer-term perspective, we assessed the memorability of a reframe, which may be essential for future utility. Nevertheless, evaluating long-term outcomes is critical and forms an important future research direction. Finally, we emphasize that our study does not investigate short-term or long-term clinical outcomes.

\section*{Acknowledgements}
\label{sec:acknowledgements}
We are grateful to the mental health practitioners and clinical psychology graduate students for data annotation, as well as the MHA visitors for participating in our field study. We thank the members of the UW Behavioral Data Science Group for their suggestions and feedback throughout the course of this project. We also thank Justin Evans for their assistance in model deployment, Xiang Lorraine Li for their input on data collection and Sebastin Santy for their input on the tool interface. T.A., A.S. and I.W.L. were supported in part by NSF grant IIS-1901386, NSF CAREER IIS-2142794, NSF grant CNS-2025022, NIH grant R01MH125179, Bill \& Melinda Gates Foundation (INV-004841), the Office of Naval Research (\#N00014-21-1-2154), a Microsoft AI for Accessibility grant, and a Garvey Institute Innovation grant.

\bibliography{_references}
\bibliographystyle{acl_natbib}

\appendix
\clearpage

\clearpage

\section{Method} \label{appendix:method}


\subsection{Linguistic Attributes of Reframed Thoughts} \label{appendix:linguistic_attributes}

We provide additional detail on the approaches described in \S \ref{subsec:metrics}.


\xhdr{Actionability} As described in \S \ref{subsec:metrics}, we measure actionability using: $contains\_action(\mathbf{R_i})$, and  $next\_action\_coherence(\mathbf{R_i})$. 

For $contains\_action(\mathbf{R_i})$, our few-shot in-context learning approach proceeds as follows. Using the reframed thoughts that were annotated as high or low actionable in our collected data (\S\ref{subsec:annotation_procedure}), we manually create 10 demonstration examples. If a reframed thought contains an action, we ask GPT-3 to extract the action from it. Otherwise, we ask it to generate the text ``\textit{No Action}''. Appendix \ref{appendix:action-prompt} shows examples. We then use these 10 demonstrations as in-context examples, followed by the reframe $\mathbf{R_i}$ which we aim to classify. If GPT-3 predicts an action for $\mathbf{R_i}$, we assign $contains\_action(\mathbf{R_i}) = 1$; else we assign 0.

For $next\_action\_coherence(\mathbf{R_i})$, we instruct GPT-3 to generate $k=5$ possible next actions given a reframed thought. Given ($\mathbf{S_i}$, $\mathbf{T_i}$, $\mathbf{R_i}$), let $\mathbf{A_i} = a_{i1}, a_{i2}, ..., a_{ik}$ be the generated set of next actions. Let $emb(\cdot)$ denote RoBERTa embeddings. Then, we define $next\_action\_coherence(\mathbf{R_i})$ as the average cosine similarity between $emb(a_i)$ and $emb(a_j)$ for all ${a_i,a_j} \in \mathbf{A_i}$.


\subsection{Action Generation Prompt}
\label{appendix:action-prompt}

We use the following prompt template for extracting actions through GPT-3:

\begin{quote}
    \textbf{Statement}: ``My bank card could be in many different places and I want to check them first before making any conclusions''	 \\
    \textbf{Proposed Action}: ``Check bank card.''
\end{quote}

\begin{quote}
    \textbf{Statement}: ``I cancelled that trip because I had to. It hurts to have done so but it was the right thing''	\\ 
    \textbf{Proposed Action}: None
\end{quote}

Also, we use the following instruction prompt for generating the next set of actions through GPT-3: ``\textit{Suggest 5 actions that the person could take based on the following statement:}''

\subsection{Hyperparameter Choices for our Proposed Retrieval-Enhanced In-Context Learning Method}
\label{appendix:hyperparameters}
For the number of examples to retrieve, we experimented with $k=1,5,10$ and $20$ and found $k=5$ to generate the most effective reframed thoughts, based on a qualitative assessment of 100 manually written situations and thoughts.

\section{Reproducibility}
Codes and datasets created in the paper will be shared at \href{https://anonymous}{https://anonymous} under an academic, attribution-only license. The use of existing artifacts was consistent with their intended use. For GPT-3 based models, we will use the OpenAI library. For other deep learning models, we train and them on two NVIDIA Titan RTX GPUs. We use the evaluate python library (\href{https://pypi.org/project/evaluate/}{pypi.org/project/evaluate}) for measuring BLEU and ROUGE scores and scipy for statistical tests. 

\section{Flagged Reframes}
\label{appendix:flagged-reframes}
There were 32 reframing suggestions out of 5,760 which were flagged (0.56\%). 19 of them were generic (59\%). 5 of them made incorrect assumptions about the person’s situation (16\%). And 8 of them may not have been relatable to the person (25\%). Importantly, we did not find any flagged reframes that were harmful or unsafe, which is critical in these scenarios. In future, exploring ways to create more personalized reframes could help avoid generic, assumptive or less relatable reframes.

\clearpage

\section{List of Thinking Traps}
\label{appendix:thinking-traps}
\begin{table}[htb!]
\resizebox{ \textwidth}{!}{
\def\arraystretch{1.2}
\begin{tabular}{@{}p{5cm}p{7cm}p{7cm}@{}}
\toprule
Thinking Traps              & Description                                                                                         & Example                                                                        \\ \midrule
All-or-Nothing Thinking     & Thinking in extremes.                                                                               & If it isn't perfect, I failed. There's no such thing as “good enough”.         \\
Overgeneralizing            & Jumping to conclusions based on one experience.                                                     & They didn't text me back. Nobody ever texts me back.                           \\
Labeling                    & Defining a person based on one action or characteristic.                                            & I said something embarrassing. I'm such a loser.                               \\
Fortune Telling             & \begin{tabular}[c]{@{}l@{}}Trying to predict the future. Focusing on \\ one possibility and ignoring \\ other, more likely outcomes.\end{tabular} & I'm late for the meeting. I'll make a fool of myself.                          \\
Mind Reading                & Assuming you know what someone else is thinking.                                                    & She didn't say hello. She must be mad at me.                                   \\
Emotional Reasoning         & Treating your feelings like facts.                                                                  & I woke up feeling anxious. I just know something bad is going to happen today. \\
Should Statements           & Setting unrealistic expectations for yourself.                                                      & I shouldn't need to ask for help. I should be independent.                     \\
Personalizing               & Taking things personally or making them about you.                                                  & He's quiet today. I wonder what I did wrong.                                   \\
Disqualifying the Positive  & \begin{tabular}[c]{@{}l@{}}When something good happens, you ignore it \\ or think it doesn't count.\end{tabular}                               & I only won because I got lucky.                                                \\
Catastrophizing             & Focusing on the worst-case scenario.                                                                & My boss asked if I had a few minutes to talk. I'm going to get fired!          \\
Comparing and Despairing    & Comparing your worst to someone else's best.                                                        & My niece's birthday party had twice the amount of people                       \\
Blaming                     & Giving away your own power to other people.                                                         & It's not my fault I yelled. You made me angry!                                 \\
Negative Feeling or Emotion & Getting “stuck” on a distressing thought, emotion, or belief.                                       & I am feeling lonely.                                                           \\ \bottomrule
\end{tabular}}
\end{table}

\clearpage

\section{Example Illustrating Our Rationality Measurement}
\label{appendix:rationality}
\begin{figure}[hbt!]
    \caption{To measure reasoning strength, we generate two explanations for each reframe -- one for why it might be sound; another for why it may be flawed. To check if the explanations themselves are well-reasoned, we recursively generate explanations for the explanations. Here, we choose a recursive tree depth of 3. Also, at every step, we generate three explanations in favour of a reframe and three explanations against it.}
    \centering
         \includegraphics[width=\textwidth]{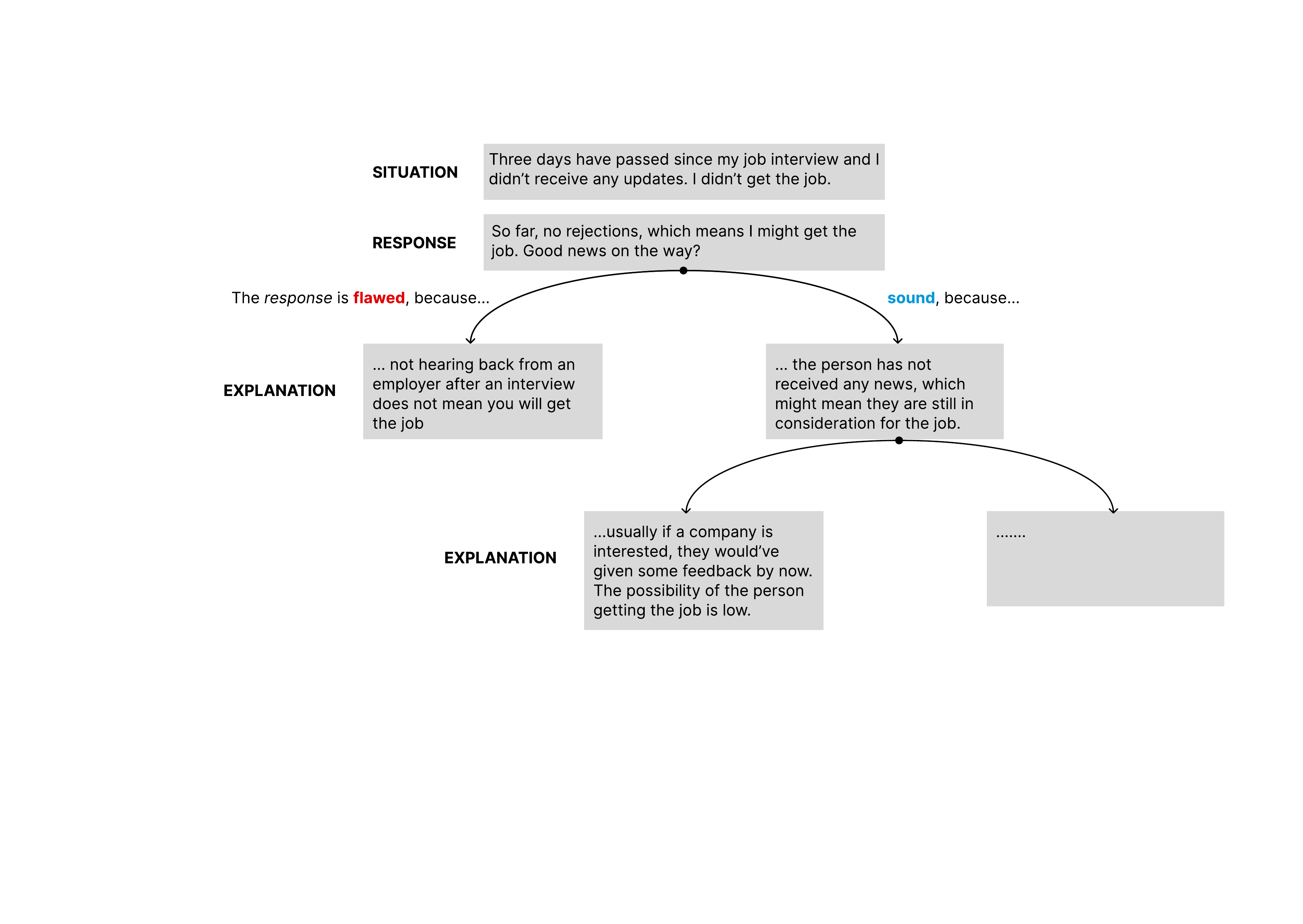}
    \label{supp:fig:rationality}
\end{figure}

\clearpage

\section{Randomized Field-Study Interface}
\label{appendix:tool-interface}
\begin{figure}[h!]
\centering
\includegraphics[width=0.9\textwidth]{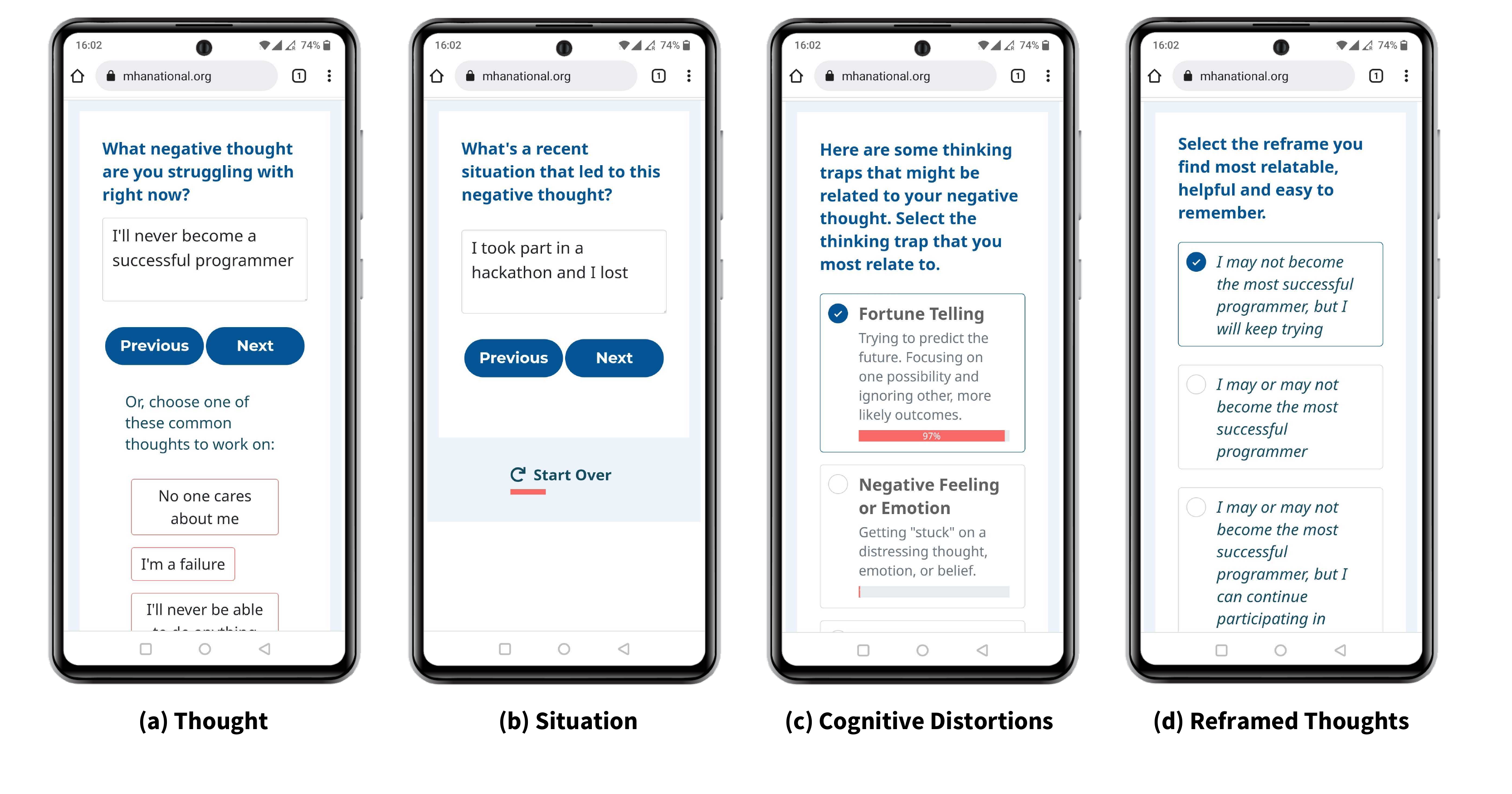}
\caption{Illustration of the interface used for our randomized field-study: (\textbf{a}) Participant starts by writing the negative thought they are struggling with in the moment; \textbf{(b)} We ask the participant to describe a recent situation that may have led to their thought; \textbf{(c)} An AI model identifies possible cognitive distortion(s) in the thought. Participant selects the cognitive distortions that they most relate to; \textbf{(d)} An AI model generates and suggests three different reframed thoughts that may help overcome negative thinking and the associated cognitive distortion. Participant selects the reframe they find the most relatable, helpful and memorable. Some of the instructions provided to the participants, including informed consent and evaluation, have been omitted from this illustration for brevity.}
\label{supp:fig:tool}
\vspace{-10pt}
\end{figure}

\clearpage

\section{Data Collection Instructions}
\label{appendix:data-collection-instructions}
\begin{figure}[hbt!]
    \caption{Instructions shown during data collection with mental health experts. Continued on the next page (1/3).}
    \centering
         \includegraphics[width=\textwidth]{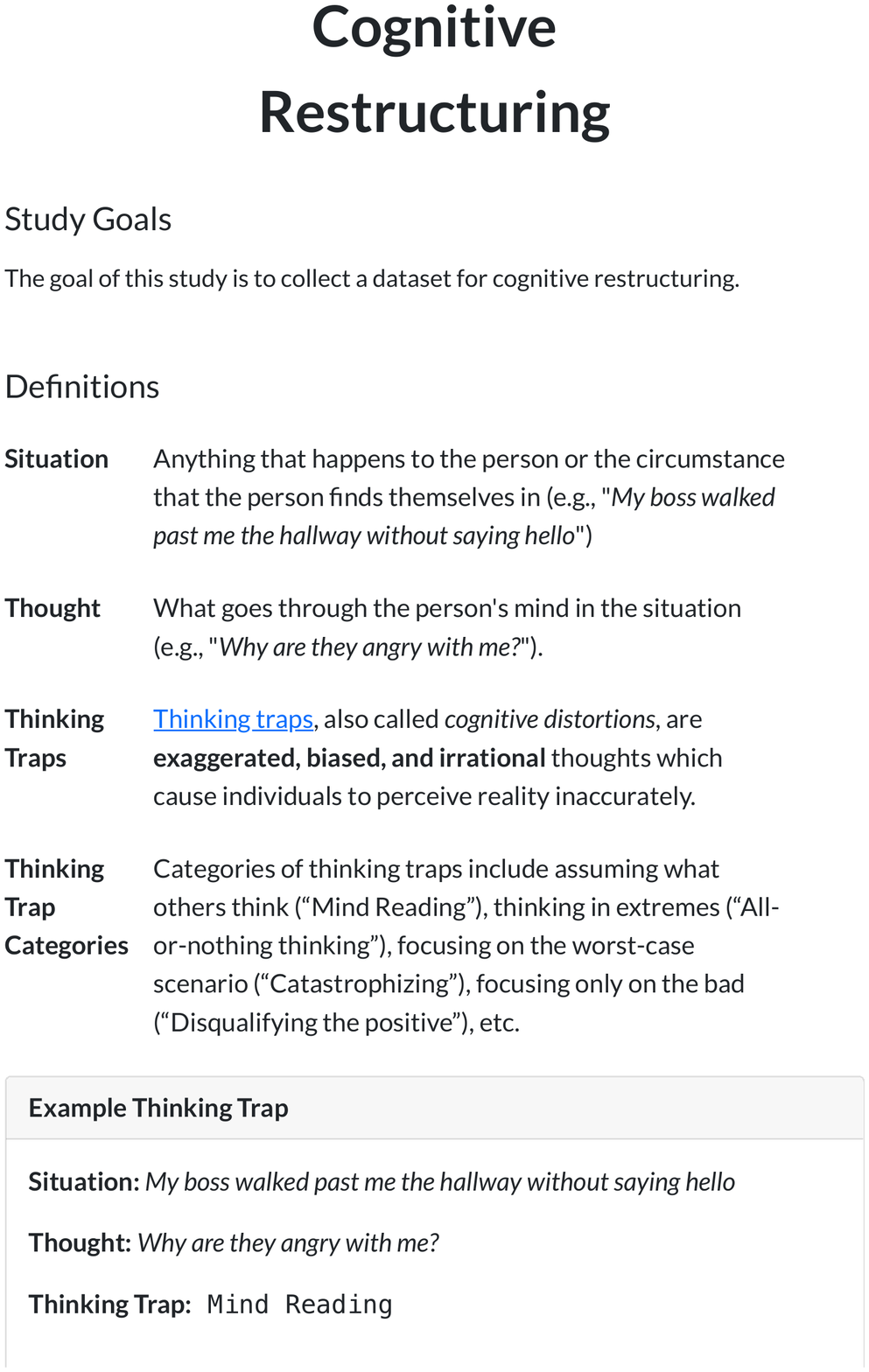}
    \label{supp:fig:data-instructions-1}
\end{figure}

\begin{figure*}[hbt!]
    \caption{Instructions shown during data collection with mental health experts. Continued on the next page (2/3).}
    \centering
         \includegraphics[width=\textwidth]{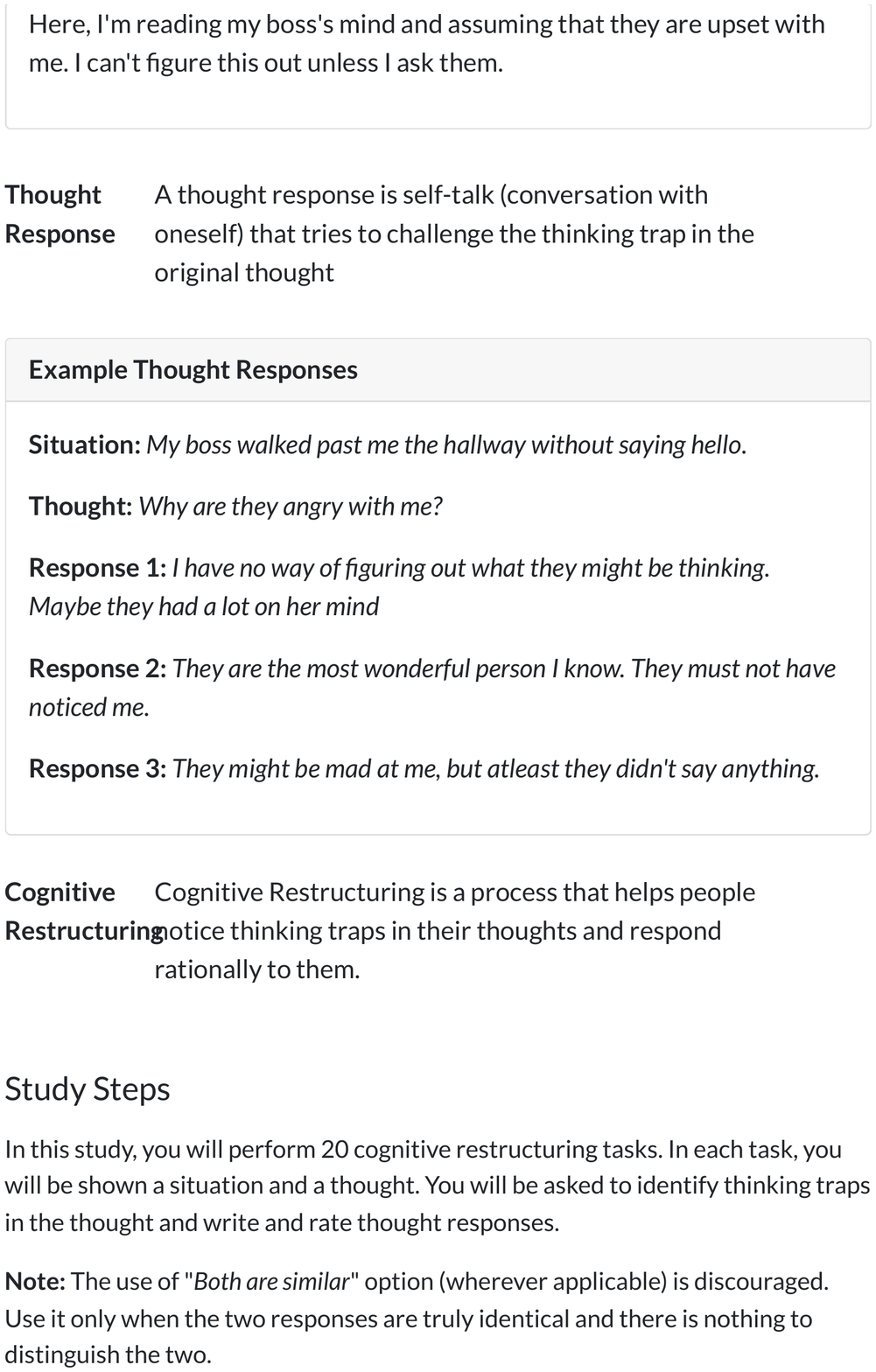}
    \label{supp:fig:data-instructions-2}
\end{figure*}

\begin{figure*}[hbt!]
    \caption{Instructions shown during data collection with mental health experts (3/3).}
    \centering
         \includegraphics[width=\textwidth]{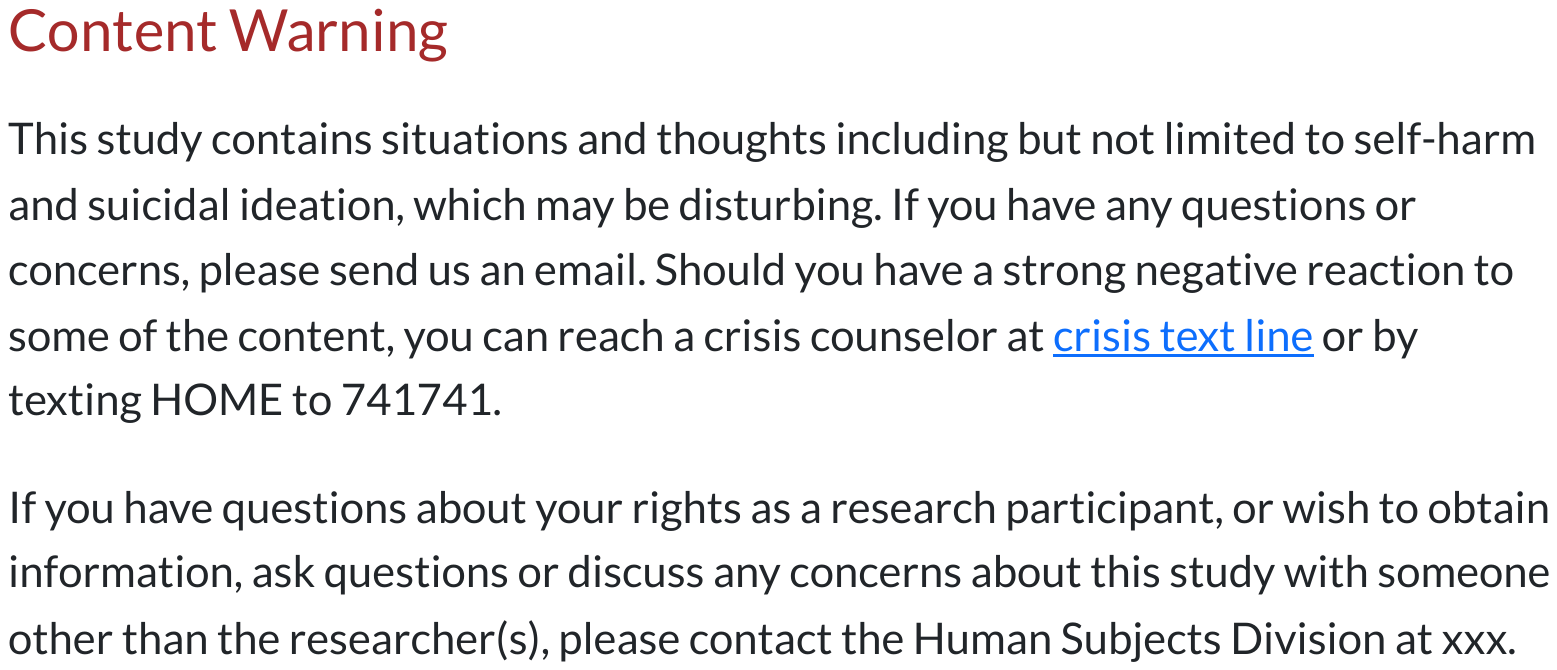}
    \label{supp:fig:data-instructions-3}
\end{figure*}

\clearpage
\section{Consent Form Used in the Randomized Field Study on MHA}
\label{appendix:consent}
\begin{figure}[hbt!]
    \caption{Consent form shown to the MHA visitors. Continued on the next page (1/2).}
    \centering
         \includegraphics[width=\textwidth]{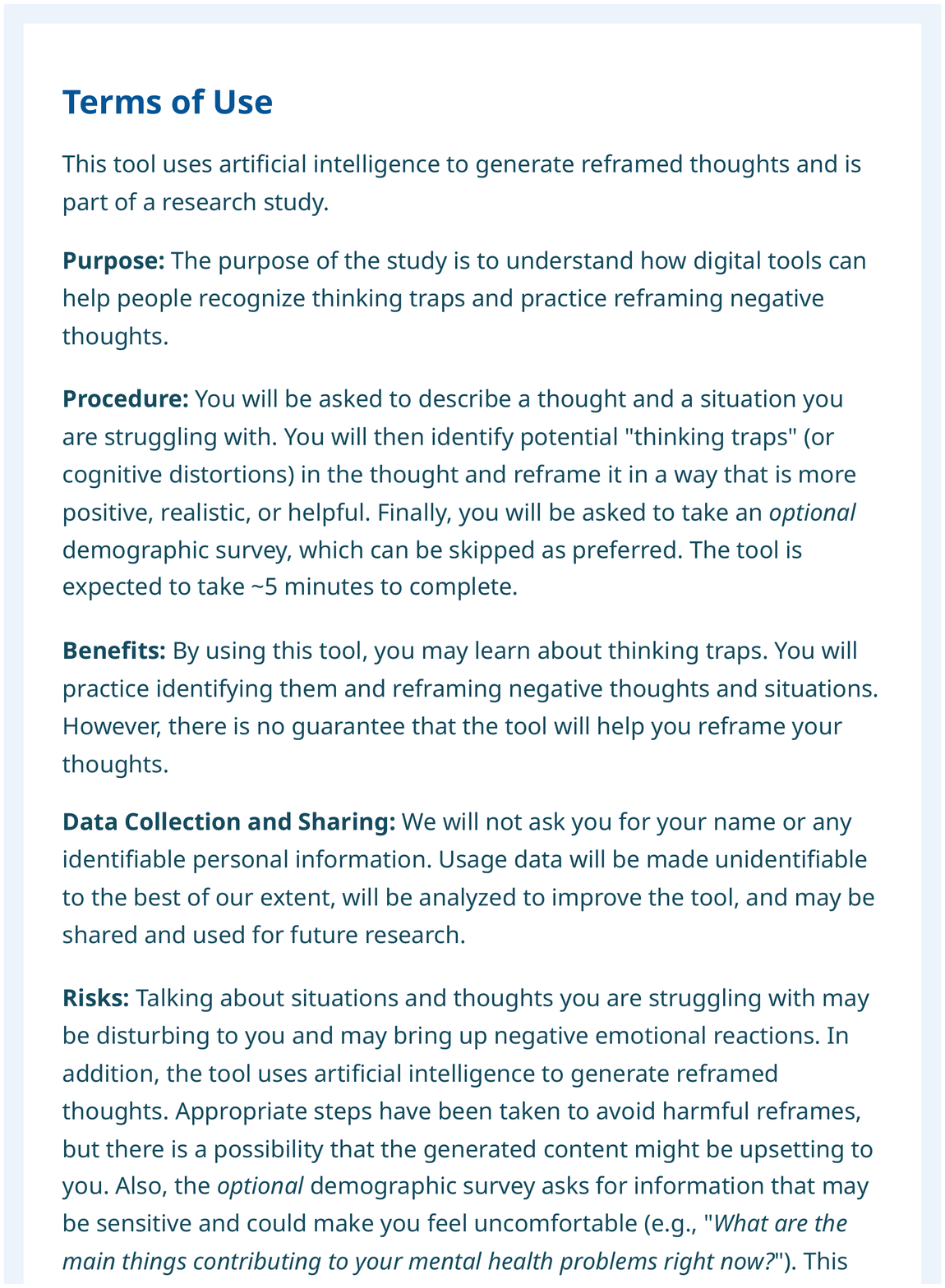}
    \label{supp:fig:consent-1}
\end{figure}

\begin{figure*}[hbt!]
    \caption{Consent form shown to the MHA visitors (2/2).}
    \centering
         \includegraphics[width=\textwidth]{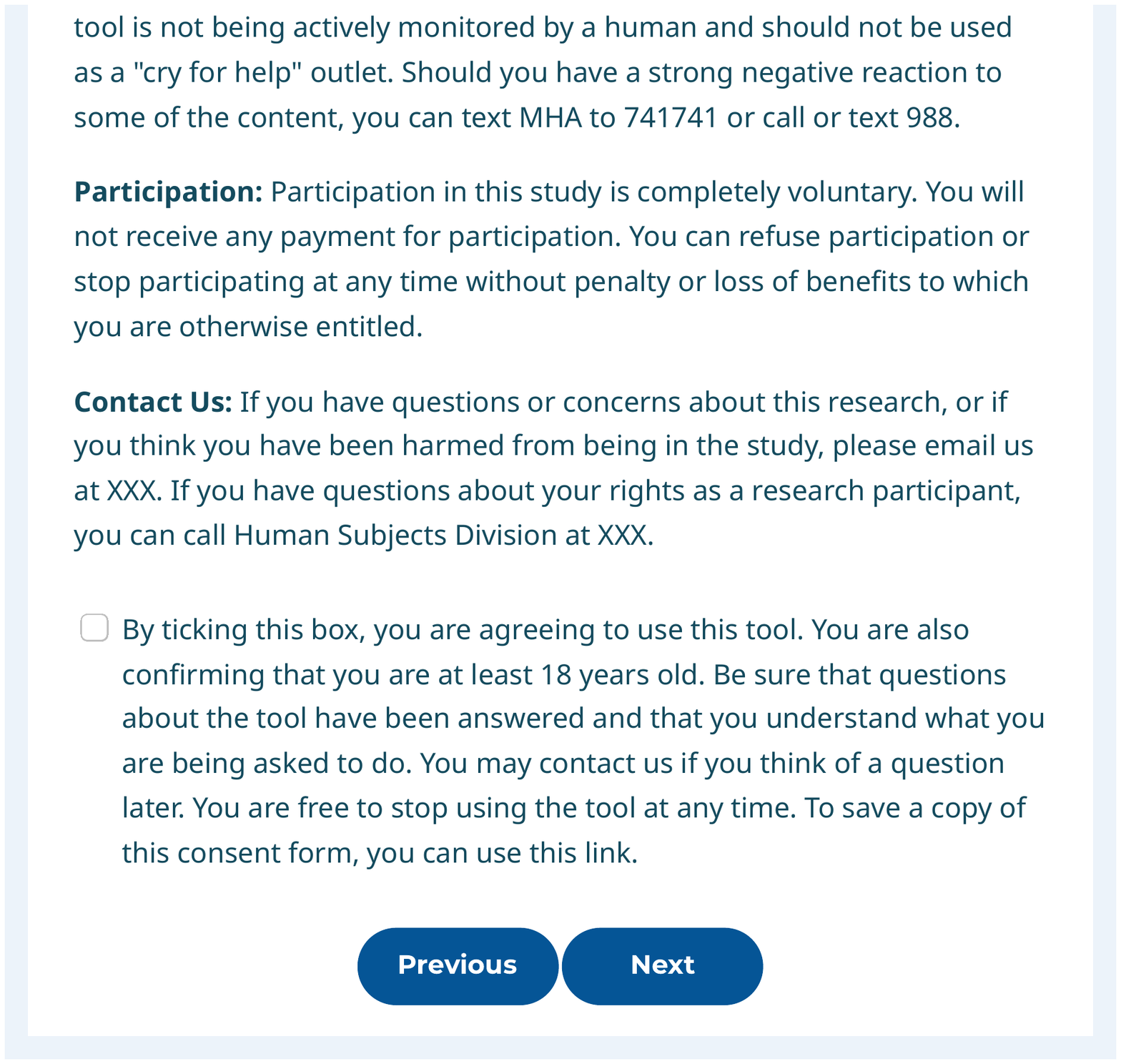}
    \label{supp:fig:consent-2}
\end{figure*}

\end{document}